\documentclass{cta-author}

\usepackage{pdfpages}
\usepackage[nolist]{acronym}

{}
{}
{}

\begin{document}


\title{Evaluating Machine Learning Models for the Fast Identification of Contingency Cases}

\author{\au{Florian Sch{\"a}fer$^{1\corr}$}, \au{Jan-Hendrik Menke$^{1,2}$}, \au{Martin Braun$^{1,2}$}}

\address{\add{1}{Department of Energy Management and Power System Operation $e^2n$, University of Kassel, 34121 Kassel, Germany}
\add{2}{Department of Grid Planning and Grid Operation, Fraunhofer IEE, 34121 Kassel, Germany}
\email{florian.schaefer@uni-kassel.de}}

\begin{abstract}
Fast approximations of power flow results are beneficial in power system planning and live operation. In planning, millions of power flow calculations are necessary if multiple years, different control strategies or contingency policies are to be considered. In live operation, grid operators must assess if grid states comply with contingency requirements in a short time. In this paper, we compare regression and classification methods to either predict multi-variable results, e.g. bus voltage magnitudes and line loadings, or binary classifications of time steps to identify critical loading situations. We test the methods on three realistic power systems based on time series in 15\,min and 5\,min resolution of one year. We compare different machine learning models, such as multilayer perceptrons (MLPs), decision trees, k-nearest neighbours, gradient boosting, and evaluate the required training time and prediction times as well as the prediction errors. We additionally determine the amount of training data needed for each method and show results, including the approximation of untrained curtailment of generation. Regarding the compared methods, we identified the MLPs as most suitable for the task. The MLP-based models can predict critical situations with an accuracy of 97-98\,\% and a very low number of false negative predictions of 0.0 - 0.64\,\%.
\end{abstract}

\maketitle



\section{Introduction}
Power flow results are the basis for power system planning and are needed in live operation to assess the system state. Quasi-static time series simulations allow evaluating asset loadings, voltage profiles and contingency situations over a long period, e.g., multiple years. This has several advantages in the planning process compared to single "worst-case" analysis including the calculation of grid losses or the integration of demand and generation flexibility \cite{Kays_2016, schaefer2019_cired}. However, the computational effort is very high. Millions of power flow calculations are necessary if multiple years, different control strategies or contingency policies (``{N-1}'' cases) are to be taken into account. For example, the simulation of one year in 15\,min resolution ($T=35,040$ time steps) for a grid with $N$ lines requires $(N + 1) \cdot T$ power flow calculations, if the single contingency policy (SCP) criterion is taken into account. In live operation, grid operators must assess if a loading situation is ``N-1'' secure in a very short time. Here, fast approximations of contingency results, including line loadings and bus voltages, are helpful to determine the system security state rapidly. A promising method to identify critical loading situations is to use \acp{ANN} as a regressor, as shown in \cite{8587482}. In this paper, we want to extend this approach to be able to use classification methods and additionally compare other regression models. It is our goal to:
\begin{itemize}
    \item identify the most suitable regression and classification methods (neural networks, ridge CV, decision trees, extra trees, random forest, gradient boosting, k-nearest neighbours) 
    \item compare the required training time and prediction time
    \item evaluate the approximation error
    \item determine the amount of needed training data
    \item test the approximation when using generation flexibility
    \item show results with two different training data sampling methods
\end{itemize}
The paper is divided into six sections. Section\,\ref{sota} gives an overview of the state of the art methods in the field of \ac{ML} in power systems and compares our approach with other publications. Section\,\ref{problem} defines the problem tackled in this work and describes how we implement the regression and classification strategy. In Section\,\ref{results}, results are shown for different \ac{ML} methods tested on three power systems. We identify the best methods, which we then compare on untrained data. In Section\,\ref{section_scenario}, we show results with an alternative training method that can be used if time series data is not available. In the last section, we give a conclusion and an outlook.
\section{State of the Art}
\label{sota}
With increasing computational power, \ac{ML} research has gained momentum in various fields \cite{Goodfellow.2016}. Comparisons in the finance sector \cite{Krauss2017} show that \acp{ANN}, gradient-boosted trees or random forests have different advantages and disadvantages depending on the problem. In power systems, \ac{ML} methods are used for many years to predict time series or contingency cases. The prediction of load \cite{Fallah2018} and generation \cite{Yang2016} time series is based on historical measurements and weather data. These methods focus solely on the time series, without taking into account the power system data, e.g. the line impedance values. The idea of contingency analysis using \ac{ML} methods was already mentioned in \cite{Maghrabi.1998}. The authors show that the prediction of bus voltages for a small test case based on a radial basis function (RBF) is possible. In this work time series are not taken into account. In comparison to modern deep learning methods, RBF neurons have a maximum activation when the center or weights are equal to the inputs. Therefore, higher extrapolation errors can be excepted. Another method to predict power flow results using \ac{ANN} is proposed in \cite{Aparaschivei2012}. By training a \ac{MLP} with P, Q bus injections, voltage magnitudes and angles are predicted. The idea is similar to the proposed regression approach. However, no contingency analysis is performed, time series are not being taken into account and results are shown only for small test systems. Different classification methods in the context of blackout predictions for a realistic test system are compared in \cite{Tomin2016}. The authors use \acp{MLP} and decision tree methods to classify system states, characterised by load level, bus voltages, power generation and contingency cases. The authors of \cite{donnot2018fast} developed a method to predict N-1 contingencies with a trained \ac{MLP}. They use a ``guided dropout'' method to generalise predictions for N-2 cases. In \cite{Canyasse_2017}, different supervised learning methods are compared to predict \ac{OPF} costs. The authors show how to reduce calculation times by using multiple regression methods, including neural networks and tree-based models. Different learning algorithms to predict real-time reliability of power systems and costs of recourse decisions are compared in \cite{Duchesne_2017}. The same authors publish an improved version of this method in \cite{Duchesne_2018}. Here, the authors show how to reduce calculation times in day-ahead operational planning, including the N-1 criterion. In \cite{Cremer_2019}, decision trees are used to learn data-driven security rules to assess and optimise power system reliability in live operation. The authors train decision tree classifier on a large number of operating points whose fault status has been determined via time-domain simulations. Power system state estimation based on \ac{ANN} is done in \cite{Menke_2019}. Here, bus voltages and line loading results are being predicted with neural networks trained with selected system states. The goal is to accurately estimate a system state in live operation with few measurements available for different switching states. Line outages and time series are not taken into account during training. A general overview of recent studies in the context of contingency analysis with artificial intelligence in planning and operational stages is given in \cite{Wu2018}. Most publications either lack of (1) realistic test grid sizes, (2) do not use grid specific time series for training and prediction, (3) do not take contingency analysis into account or (4) analyse only one \ac{ML} method. In this paper, we want to identify which approach (classification or regression) and which models are best adapted for similar problems.

\section{Implementation of the Regression and Classification Methods}
\label{problem}\label{implementation}
\begin{figure}[!ht]
\centering
\includegraphics[width=0.45\textwidth]{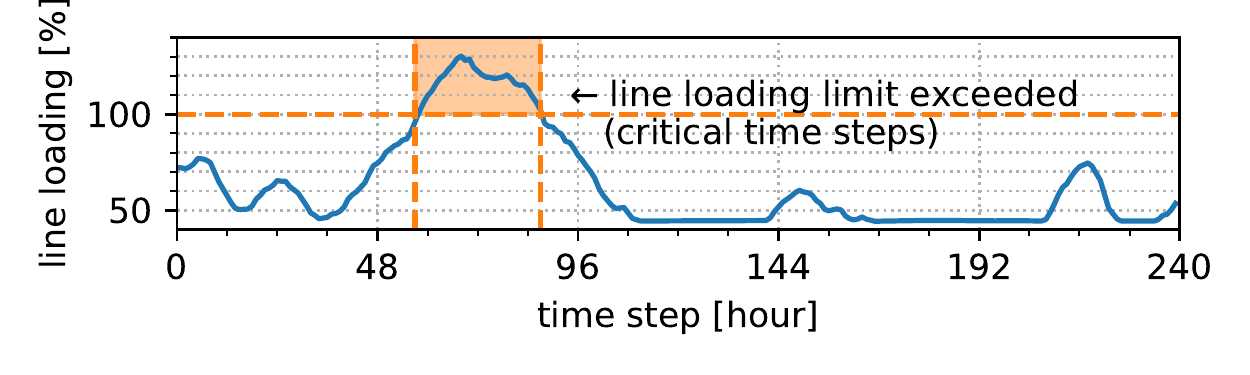}
\caption{Example of critical time steps resulting due to high line loadings.}
\label{fig_critical_steps}
\end{figure}

We use \ac{ML} model implementations from \cite{scikit-learn, Chen:2016:XST:2939672.2939785} to identify time steps with high line loadings or voltage violations for the given time series and grid data. The objective is to significantly reduce the calculation time with a minimal loss in precision by training either a regressor or classifier with a certain percentage of time steps of the power flow results and corresponding inputs. We use the regression and classification method to predict important system variables. The regressor is trained to predict the voltage magnitudes $V_m$ of all buses and line loading values $I_{\%}$ of all lines in the grid for a time step. By comparing with the pre-defined limits, it can then be assessed if the time step is critical. The classifier is directly trained to predict whether a time step is critical. Figure \ref{fig_critical_steps} shows exemplary line loading results for 10 consecutive days with exceeded line loading limits (critical) between time step 57 and 87.
\subsection{Methods Overview}
\begin{figure}[!ht]
\centering
\includegraphics[width=0.49\textwidth]{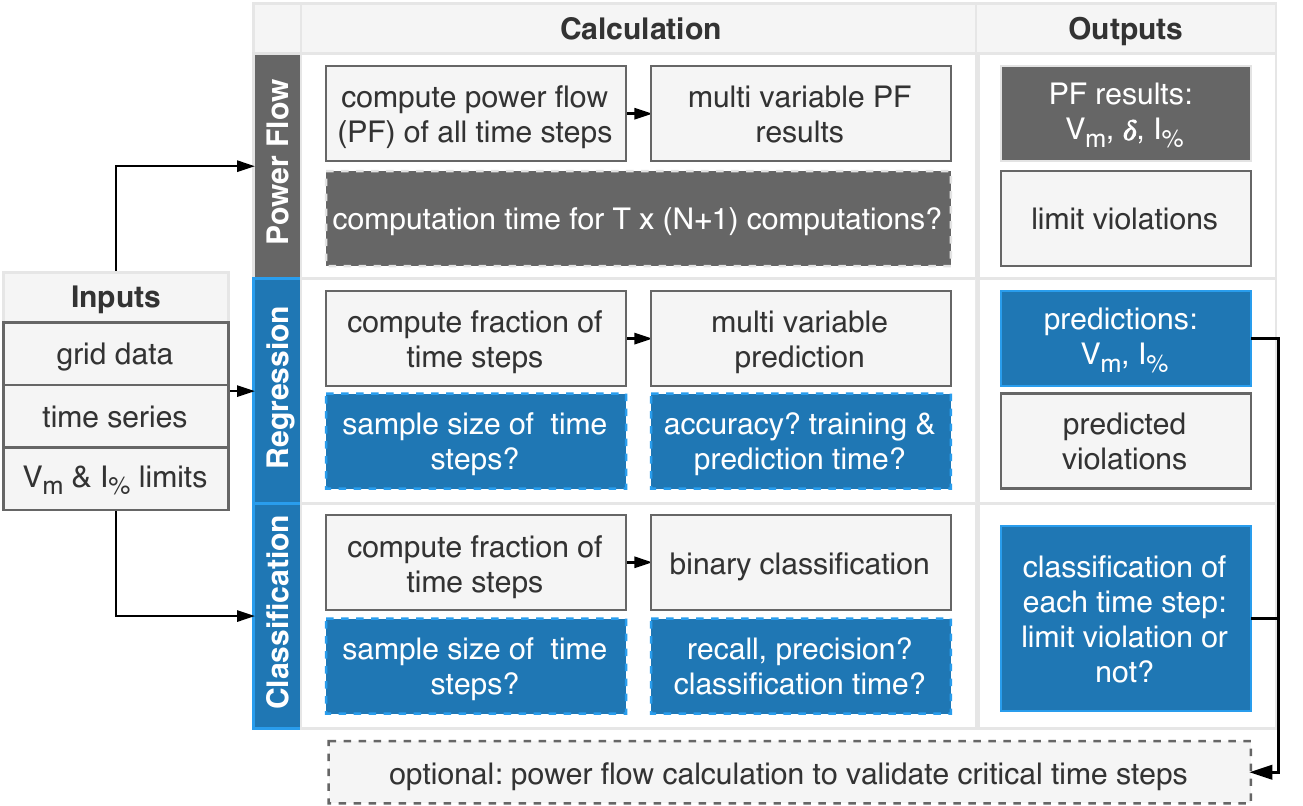}
\caption{Overview of the quasi-static power flow, regression and classification method including research questions (dashed boxes).}
\label{fig_sim}
\end{figure}

Fig.~\ref{fig_sim} shows an overview of the methods. The input data is identical for all methods and consists of the grid data with a fixed topology (switching state) and the real power $P$ and reactive power $Q$ time series for loads and \ac{RES} (PQ-nodes). We model large generation units as PV-nodes. For these generators, real power injections and bus voltage magnitudes are varied. In the following comparison, we assume static voltage set-points for PV- and slack-nodes. The time series can be derived from historical measurement data or by simulation. The power flow method iterates over all time steps, updates the $P, Q$ values, and the bus voltage magnitudes $V_m$, voltage angles $\delta$ and branch currents $I_\%$. We compute line outages for each line $l \in N$, which results in $N$ additional power flow calculations for each time step $t \in T$. In total $(N + 1) \cdot T$ power flow calculations must be calculated to obtain results for the base case and all contingency results. This process can take up to several hours to complete.

To reduce calculation times, we simulate only a fraction of time steps to generate the training data for the regression and classification training by quasi-static power flow simulations of one year. These power flow results are the input for the regression or classification training. All regression and classification methods belong to the category of supervised learning algorithms, i.e., we provide training input data $X$ and corresponding output data $y$. The training inputs contain parts of the known inputs of a power flow calculation. Outputs for the regression methods are the voltage magnitude approximations at each bus $V_m$ and the line loading $I_\%$ of each line in per cent of the maximum line loading. Outputs for the classification methods are ``critical'' system state $s=1$ or ``uncritical system state'' $s=-1$. We define a critical system by the violation of voltage magnitude limits or line loading limits. Optionally, power flow results can then be computed for the classified critical time steps or to validate regressor predictions. To test the proposed methods, we use time series data of one year with and without curtailed power injection of generators. Validation criteria for the regression method are the mean and the maximum error. We measure classification success by standard criteria, such as precision, recall and accuracy.

\subsection{Input Data}
\label{input_data}
\subsubsection{Input Layer and Architecture}
\label{input_layer}
The input layer is identical for the regressor and classifier. Each feature, defined by \eqref{eq:X}, contains parts of the known variables of the power flow calculation for a time step $t$: 

\begin{equation}
X_t =
\begin{bmatrix}
v_{m,r} & \delta_{r} & v_{m,gen} & p_{bus} & q_{bus}\label{eq:X} 
\end{bmatrix} \\
\end{equation}

with $v_{m,r}$ and $\delta_{r}$ the voltage magnitude and angle of the reference buses and $v_{m,gen}$ the voltage magnitudes of the generator buses (PV-nodes). $p_{bus}$ are the sum of the known real power values per bus including all loads and generators. $q_{bus}$ are the known reactive power values of PQ-buses (aggregated load and \ac{RES} values). We use the default hyperparameter settings of \cite{scikit-learn}, which show good results for common problems. A separated model is trained for each N-1 case.
%

\subsubsection{Training Data}
\label{time_series_simulation}
We compute power flow results with pandapower \cite{Thurner.2018} by iterating over all time steps. In the following comparisons, we include line contingency cases by setting each line out of service one after another and calculating the power flow results for the whole year. Depending on the total number of time steps $T$ and the number of lines $N$ in the grid, this process takes $T \cdot (N+1) \cdot t_{pf}$ seconds, where $t_{pf}$ is the average time for a single power flow calculation. We split the same time series in a training and a prediction set. An alternative method to generate training data is to use the scenario generator from \cite{Menke_2019} or vine copulas as published in \cite{Konstantelos_2019}. The creation of training data with these methods is especially useful if no time series data is available or to obtain additional data for future planning. In Section\,\ref{section_scenario}, we show prediction results when using the scenario generator. The training with the scenario generator outlines that the model architecture is able to generalise from the training data and that it does not only learn to predict the remaining part of the time series. 

\subsection{Regression Method}

An individual regressor model is trained to predict voltage magnitudes $v_{m, \mathrm{bus}}$ in per unit (p.u.) values of all buses as well as the line loadings $I_{\%, \mathrm{line}}$ in percent of the rated current $I_{r}$. 


\label{regression}

\begin{equation}
y_1 =
\begin{bmatrix}
v_{m,\mathrm{bus}}
\end{bmatrix} \\
y_2 =
\begin{bmatrix}
I_{\%,\mathrm{line}} \label{eq:y1y2_reg}
\end{bmatrix} \\
\end{equation}

Relevant performance metrics for the regression method are the mean absolute error of the predictions as well as the maximum error. A low mean error is relevant if the regressor is used to predict the results of similar time steps. A low maximum error is needed when critical time steps/loading situations are to be identified.

\subsection{Classification Method}
\label{classification}

The classification output layer is defined by a binary state where $s=-1$ equals "uncritical" time step and $s=1$ equals "critical" time step:
\begin{equation}
y_\mathrm{classifier} =
\begin{bmatrix}
-1 & 1 \\
\end{bmatrix} \\
\label{eq_inputs}
\end{equation}

A time step is critical when either the voltage magnitude of any bus is out of boundaries $v_m < v_{\mathrm{min}}$, $v_m > v_{\mathrm{max}}$ or the line loading $I_{\%}$ of at least one line violates its maximum ($I_{\%} > I_{\mathrm{limit}}$). Operational restrictions for specific grids are defined by the power system operator individually. Typically, they are derived from standards such as the VDE-AR-N~4121~\cite{VDE_4121}. We define a critical system state if, for any bus in the grid, the voltage magnitude violates a range between $0.9\,p.u.$ - $1.1\,p.u.$ of the nominal voltage magnitude $V_m$ or the long term thermal line loadings is above their maximum loading $I_{\mathrm{limit}}$.

\subsubsection{Performance Metrics}
\label{metrics}
The classifier should preferably predict uncritical time steps as critical (false negatives) and be less precise than fail to notice critical time steps (false positives). Different metrics are commonly used to assess the performance of classifiers. Recall \eqref{eq:recall} measures the fraction of true positive (TP) classifications over the total amount of relevant instances. The relevant instances are the sum of TP and false negative (FN) classifications. Here, TPs are the correctly identified critical time steps and FN are critical time steps which have been mislabelled as uncritical. 
\begin{align}
\text{recall} = \frac{TP}{TP + FN}\label{eq:recall}
\end{align}

In our case, the FN classifications should be minimised, since it is of high importance to identify all critical loading situations. We preferably tolerate some uncritical time steps identified as critical (false positives (FP)) than a high recall. The precision score measures the misclassification:
\begin{equation}
\text{precision} = \frac{TP}{TP + FP}\label{eq:precision}.
\end{equation}

When classifying critical loading situations, maximising recall is more important than maximising precision. The accuracy score \eqref{eq:accuracy} measures the ratio of correct classifications to all classifications. Here, TNs are the true negatives, which are the correctly classified uncritical time steps. The accuracy metric alone can be misleading for imbalanced datasets, where the majority of time steps are uncritical, and only a fraction is critical. In this case, the accuracy score is high by default when labelling every time step as uncritical.
\begin{equation}
\text{accuracy} = \frac{TP + TN}{TP + TN + FP + FN}\label{eq:accuracy} 
\end{equation}


\subsubsection{Training of Imbalanced Datasets}
The training data is imbalanced since the majority of time steps is "uncritical" with a few critical time steps to be identified (compare Fig.\,\ref{fig_critical_steps}). We can achieve a high accuracy if all of the time steps are labelled as "uncritical". However, recall is also small in this case since the number of false negative is maximal. To overcome this issue, we predict the probabilities to which class the time step belongs to instead of predicting if the time step is critical or not. Since we have a binary classification, the classifier outputs a probability matrix of dimension $(T,2)$. The first index refers to the probability that the time step is "uncritical", and the second refers to the probability that the time step is "critical". By reducing the probability threshold for the "critical" class, the number of positive predictions and recall increase while precision decreases. Additionally, we use \ac{SMOTE} \cite{Chawla2002} as an oversampling strategy to balance the dataset.

\section{Results}
\label{results}

\subsection{Case Data}
\begin{figure}[!ht]
\centering
\includegraphics[width=0.45\textwidth]{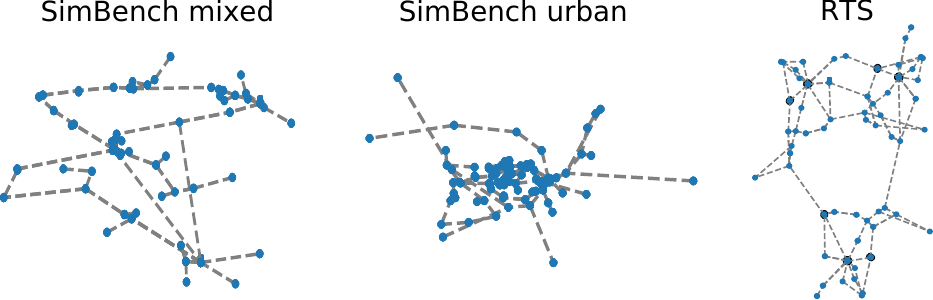}
\caption{Analysed test grids: SimBench (SB) mixed, SimBench urban, and RTS.}
\label{fig_grid}
\end{figure}

We apply the regression and classification methods on three different synthetic grid models, which are derived from real power systems. All models and the corresponding time series are available in the open-data pandapower format. 
The characteristics of the SimBench (SB) grids \cite{simbench} are typical for German meshed high-voltage grid topologies. Time series are available in $15$\,min resolution with $35,136$ time steps in total. The Reliability Test System (RTS) test case \cite{rts} is a North American power system model with a time series resolution of $5$\,min, resulting in $105,408$ time steps. In total, over 12 million power flow results must be calculated for the given time series of one year when assessing all N-1 cases. Fig.~\ref{fig_grid} shows the synthetic grids. Table~\ref{grid_data} lists the relevant data of the three test cases. We compute all results on an Intel Core i7-8700K CPU at 3.70GHz speed. 
\begin{table}[!ht]
\processtable{Overview of grid data.\label{grid_data}}
{\begin{tabular*}{20pc}{@{\extracolsep{\fill}}l|lll@{}}\toprule
& SB mixed \cite{simbench}& SB urban \cite{simbench} & RTS \cite{rts} \\
\midrule
voltage level [kV] & 110 & 110 & 230\\
buses [\#] & 64 & 82 & 73\\
N-1 cases [\#] & 66 & 78 & 66\\
$I_{\mathrm{limit}}$ [\%] & 60. & 60. & 100. \\
$N_{\mathrm{time steps}}$ & 35,136 & 35,136 & 105,408 \\
$N_{PF}$ [$10^6$] & 2.312 & 2.733 & 6.957 \\
\botrule
\end{tabular*}}{}
\end{table}

\subsection{Regression Results}
We evaluate the performance of different regression methods by comparing the absolute error of line loading and voltage magnitude predictions while taking into account the training size as well as training and prediction time. We analyse regressors from \cite{scikit-learn} that support multivariable outputs:
MLPRegressor (MLP),
ExtraTreesRegressor (ET), 
DecisionTreeRegressor (DT), 
RandomForestRegressor (RF), 
RidgeCV (RCV).

\begin{figure}[!ht]
\centering
\includegraphics[width=0.49\textwidth]{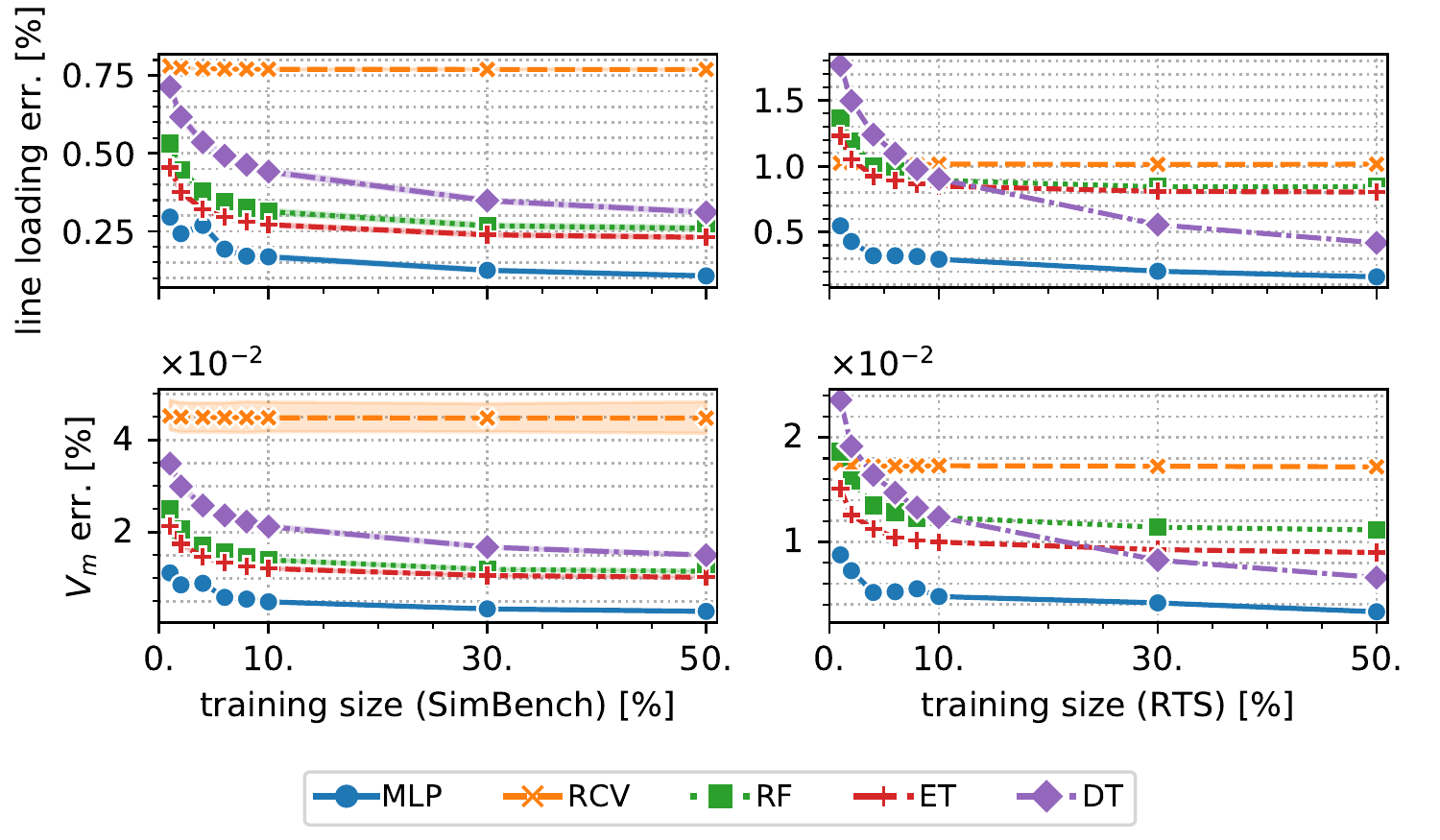}
\caption{Mean prediction error with increasing training size for both SimBench test cases combined (left) and the RTS case (right).}
\label{regression_results_training_size}
\end{figure}

First, we analyse how the prediction errors decrease with training data size. From the $31,536$ (SB) or $105,048$ (RTS) time steps, we randomly select training and test data by a shuffled train/test split. Based on the test data set, we evaluate the absolute prediction error. Fig.\,\ref{regression_results_training_size} shows the mean prediction error for the SB test cases (left) and the RTS test case (right) with increasing training sizes. All regressors improve with larger training sizes, except the RCV method. The prediction error decreases significantly with training sizes up to $10$\,\%. Larger training sizes reduce the prediction error primarily for the \ac{MLP} and DT regressors. We, therefore, use a train/test split of 0.1\,/\,0.9 for the following comparisons. 

\begin{figure}[!ht]
\centering
\includegraphics[width=0.49\textwidth]{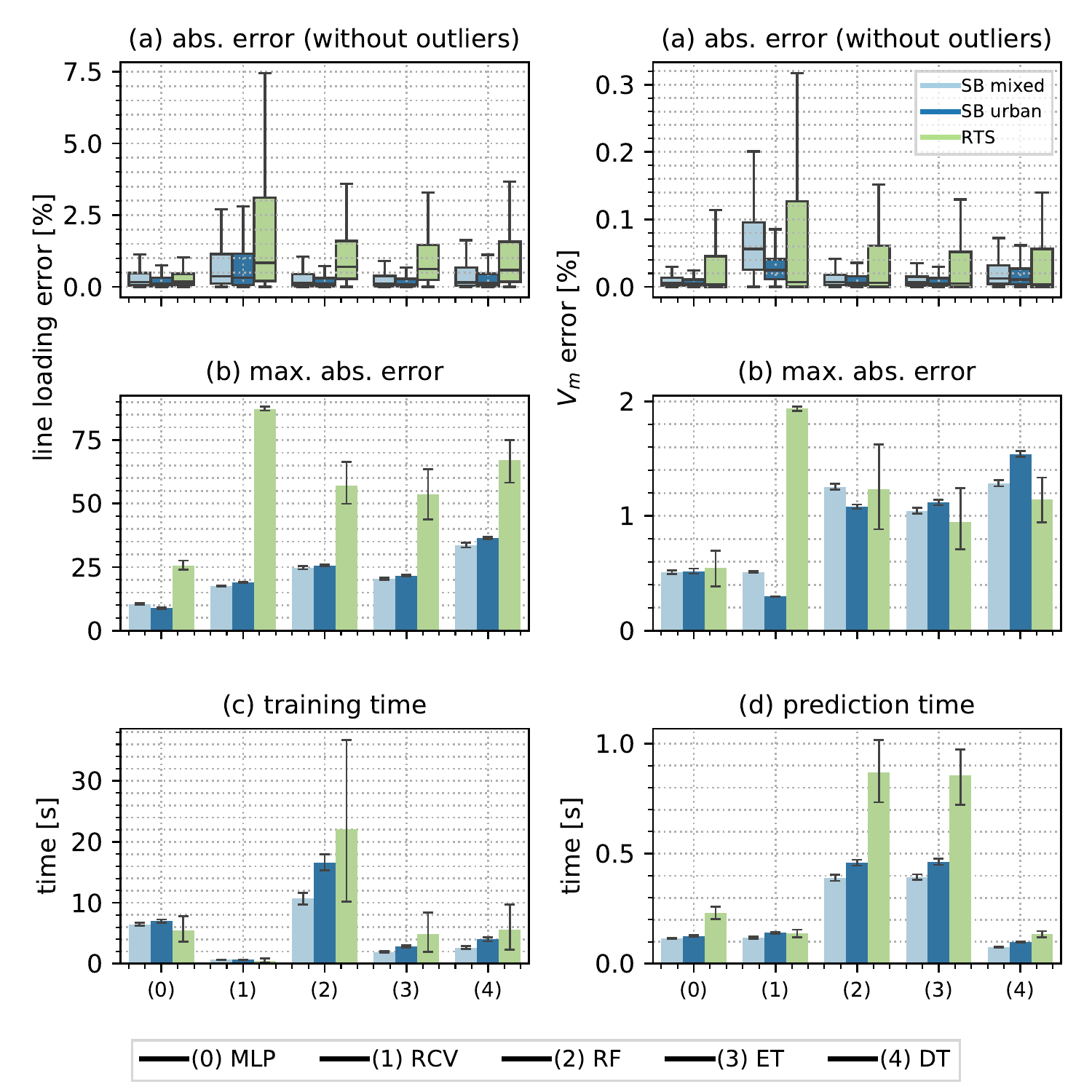}
\caption{Results for the different regressors and test cases. The whiskers show the 95\,\% percentiles.}
\label{regression_results_compare}
\end{figure}

Fig.\,\ref{regression_results_compare} (a) shows the results of the prediction error without outliers. The \ac{MLP}, ET and RF regressors yield the lowest mean error values for voltage and line loading predictions. The mean error of the regressor with the lowest error, the \ac{MLP}, is only a third in comparison to the regressor with the highest error of the RCV method. The \ac{MLP} has the lowest errors of all regressors when comparing the maximum error in Fig.\,\ref{regression_results_compare} (b). The RCV, tree, and RF methods have significantly higher prediction errors. A longer training time (Fig.\,\ref{regression_results_compare} (c)) is needed for the \ac{MLP} in comparison to RCV, ET and DT. The time needed to predict the results (Fig.\,\ref{regression_results_compare} (d)) is shortest for the DT, \ac{MLP} and RCV methods.

The \ac{MLP} has the lowest overall error with decent training and low prediction time. With $10$\,\% of all time steps being trained, the mean errors of line loading predictions are less than $0.25$\,\% for the SB grids and less than $2$\,\% for $99$\,\% of the predicted values. Similarly, the voltage magnitude prediction errors are low with a mean value of $0.01$\,\%. and $0.5$\,\%. in the $99$\,\% range. The mean prediction error for line loadings for the RTS test case is $0.5$\,\% with $99$\,\% of all values being predicted with an error less than $5$\,\%. Voltage magnitude prediction errors are $0.05$\,\%. (mean) and $0.5$\,\%. ($99$\,\%). However, some outliers cannot be predicted with this accuracy. 

\subsection{Classification Results}
The goal of the classification is to detect time steps in the data set with high line loadings or voltage tolerance violations, which are categorised in "critical" and "uncritical". Power flow results for these time steps can be calculated separately if needed. The classifiers we analyse are:
xgboost XGBClassifier (XGB) from \cite{Chen:2016:XST:2939672.2939785}, 
RandomForestClassifier (RF), 
AdaBoostClassifier (AB), 
GaussianNaiveBayes (GNB), 
ExtraTreesClassifier (ET), 
MLPClassifier (MLP), 
KNeighborsClassifier (KN) from \cite{scikit-learn}.

\begin{figure}[!ht]
\centering
\includegraphics[width=0.49\textwidth]{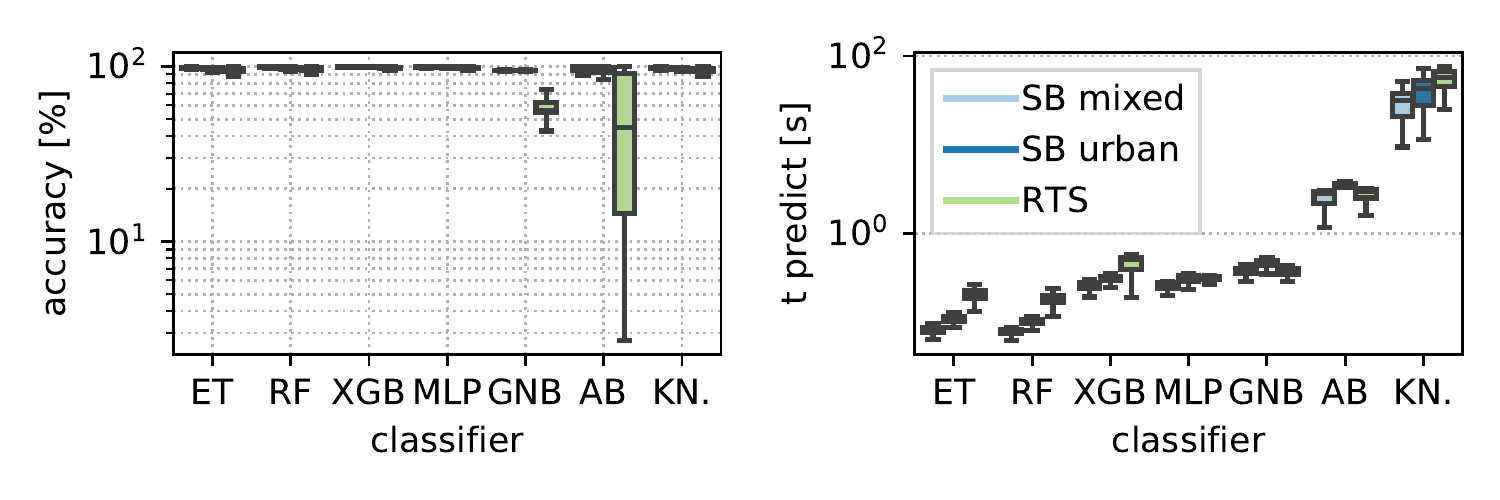}
\caption{Classification accuracy and timings for all classifiers and test cases including contingency analysis.}
\label{accuracy}
\end{figure}

Fig.\,\ref{accuracy} shows the classification accuracy and prediction timings for all classifiers. The AB and GNB classifiers have - on average - a much lower accuracy compared to the other classifiers. Their percentage of correct predictions was less than 90\,\% in all test cases. The prediction by the KN classifier takes between 30\,s and 1\,min on average in comparison to less than 0.5\,s by the other classifiers without being more accurate than the ET, RF, XGB and XLB classifier. We, therefore, conclude that the AB, GNB and KN classifiers are not as suitable for the classification of critical time steps as the other classifiers and exclude them from further comparisons.

\begin{figure}[!ht]
\centering
\includegraphics[width=0.49\textwidth]{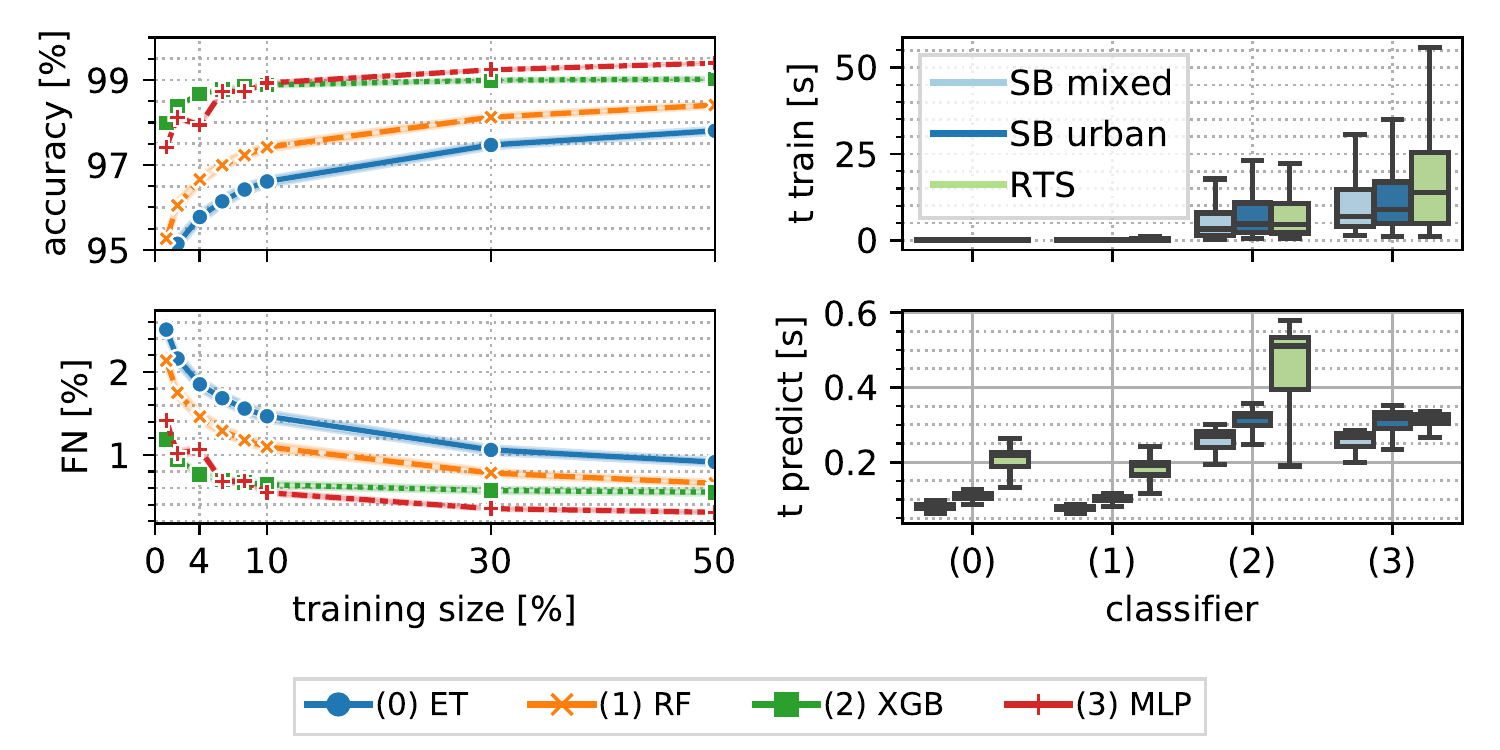}
\caption{Classification accuracy and false negative rate with increasing training size for all test cases with {N-1} predictions (left). Training and prediction times for each classifier and test case.}
\label{training_size_times}
\end{figure}

Fig.\,\ref{training_size_times} (left) shows the increasing accuracy and decreasing number of false negative predictions with increasing training size for the classifiers with an accuracy of more than 90\,\%. The ET and RF method can classify about 96 - 98\,\% of time steps correctly with approximately 1.0-1.5\,\% being false negatives. Both classifiers are outperformed by the \ac{MLP} and the XGB methods. These methods have an accuracy starting at 98\,\% at a training size of 1\,\% of all time steps being trained. With an increasing training size both, \ac{MLP} and XGB, achieve an accuracy of 99.5\,\% with 0.3\,\% of FN classifications at a training size of 50\,\% of time steps and {N-1} cases being calculated. 

Corresponding training and prediction times are shown for each grid and classifier in Fig.\,\ref{training_size_times} (right). The training time for the ET and RF classifiers is on average much shorter ($<$~0.7\,s with 50\,\% training data) compared to the \ac{MLP} and XGB methods. Depending on the training size, the \ac{MLP} training time takes $\sim$ 2.5\,s for 1\,\% of the data up to more than one minute for the RTS test case. In comparison, the XBG is twice as fast in the RTS case with 0.9\,s and 35\,s respectively. Prediction times of the ET and RF methods are about one third compared to the times needed by the \ac{MLP} and XGB methods. The time needed to predict the classification results is on average similar for the \ac{MLP} and XGB methods with an exception in the RTS case. Here, the XGB needs twice the time ($\sim$0.5\,s) of the \ac{MLP} ($\sim$0.25\,s). The difference in prediction time is negligible when taking into account the time needed to compute the training data (see Section\,\ref{section_timings}).

\begin{figure}[!ht]
\centering
\includegraphics[width=0.45\textwidth]{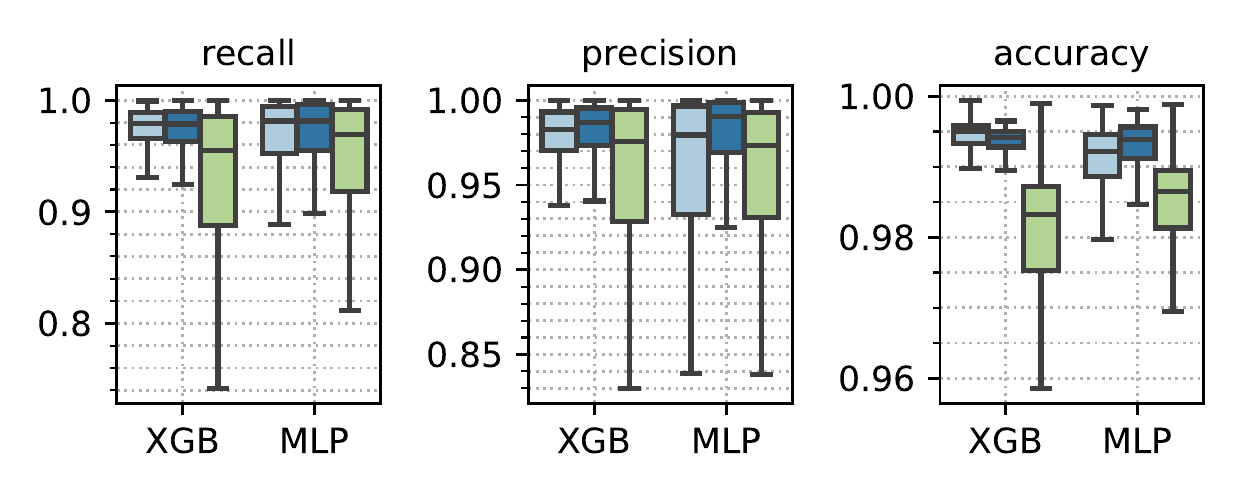}
\caption{MLP and XGB recall, precision and accuracy scores including {N-1} predictions (SB mixed, SB urban (blue), RTS (green)). We choose a prediction threshold of 0.2 with a train/test split of 0.1\,/\,0.9.}
\label{classification_results}
\end{figure}

Fig.\,\ref{classification_results} shows the recall, precision and accuracy scores for the XGB and MLP classifier with a training size of 10\,\%. Both methods yield good results with accuracy values of at least 98\,\% correct classifications on average. The average recall score of the MLP is higher compared to XGB but also has more outliers in some {N-1} case predictions. The XGB classifier achieves a higher precision, on the other hand. Therefore, the classification with the MLP and XGB may help to identify critical time steps to run detailed analysis based on power flow calculations.

\begin{figure}[!ht]
\centering
\includegraphics[width=0.49\textwidth]{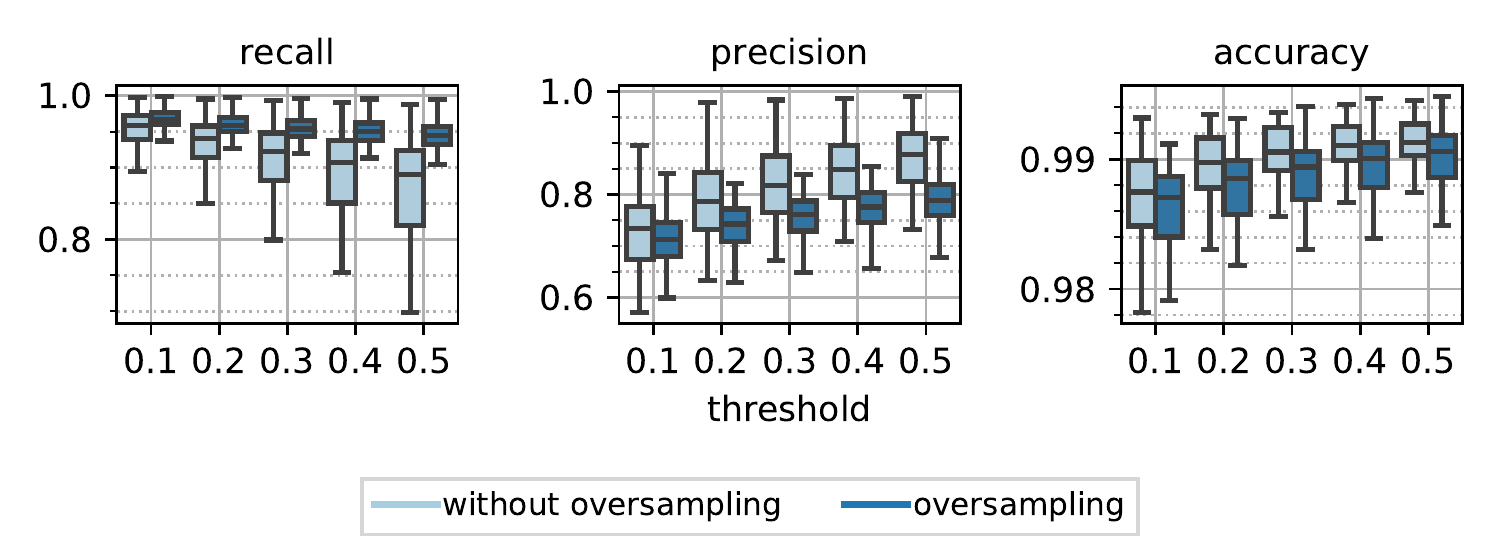}
\caption{MLP training with and without oversampling.}
\label{fig_oversampling}
\end{figure}

The training data is very imbalanced since only a few time steps are critical. Oversampling techniques, such as SMOTE \cite{Chawla2002}, allow balancing the training data. SMOTE creates additional data for training by interpolating between existing samples, and the obtained artificial data-set is then used for training. Fig.\,\ref{fig_oversampling} shows the difference in recall, precision and accuracy for the MLP classifier when using over-sampled data. Each box-plot contains the classification results of all N-1 cases and grids combined. Recall increases when using oversampling for all thresholds - similarly, the accuracy and precision decrease as expected. 
%
The absolute number of FN predictions decrease by 10.3\,\% (SB mixed), 33.01\,\% (SB urban), and 46.1\,\% (RTS) for a prediction threshold of 0.2. However, the number of FP predictions increase by 10.77\,\%, 23.7\,\%, and 27.1\,\% respectively. Note that for each FP prediction, an additional power flow calculation for verification is needed.
%
%
\subsection{Comparison of MLP Regression and Classification}
\label{comparison}
\begin{figure}[!ht]
\centering
\includegraphics[width=0.49\textwidth]{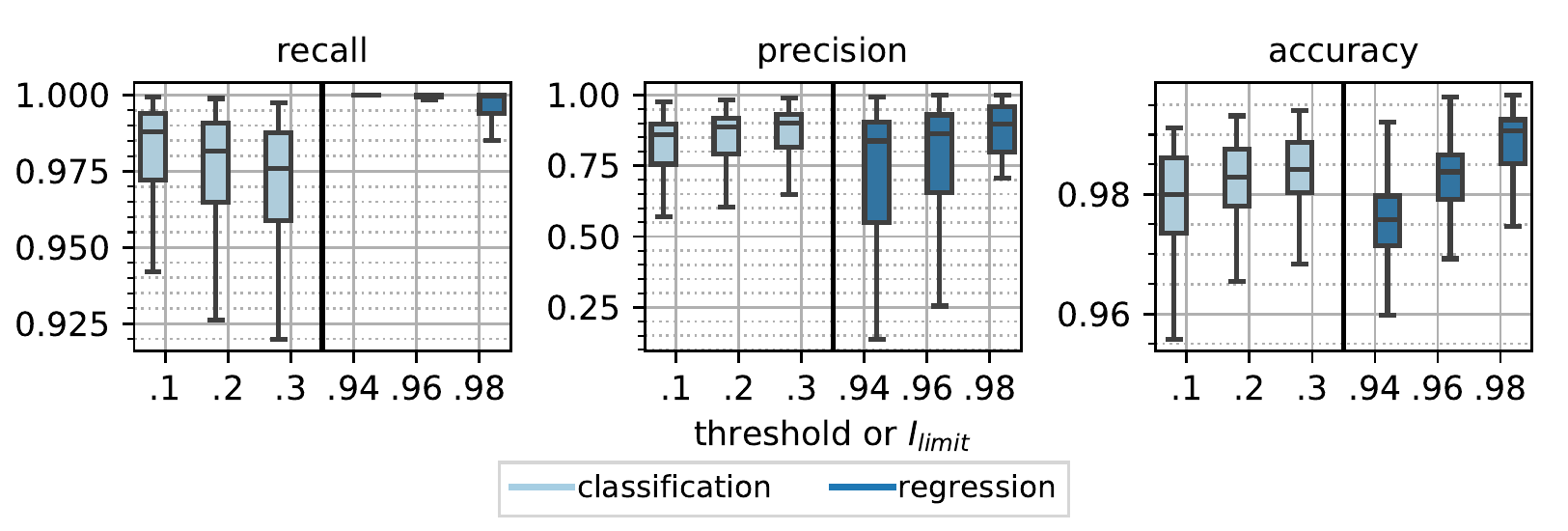}
\caption{Comparison of classification and regression results for the MLP models.}
\label{regressor_vs_classifier}
\end{figure}

To compare the classification and regression method, we categorise the results of the regression method. We define a time step as critical for each (predicted) line loading value above a threshold of the max. loading limit $I_{\mathrm{limit}}$. This is similar to setting a lower threshold for the classification. Fig.\,\ref{regressor_vs_classifier} shows the direct comparison of the MLP regression and classification methods in terms of recall, precision and accuracy. We use the oversampling method for the classification, since it showed the highest recall values. The data is obtained by a random train-test split of the time series of one year. The trained regressor has a much higher recall than the classifier, even when we use oversampling for the classifier. Recall of the regressor is close to one when setting the prediction limit to a value of $0.94 \cdot I_{\mathrm{limit}}$ for the three test cases. At this threshold, nearly no false negative predictions are made and more than 99\,\% of critical time steps are identified correctly. However, precision drops to low values in that case and accuracy decreases to mean values of less than 0.98. The precision and accuracy of the regressor significantly increase when setting a $0.98 \cdot I_{\mathrm{limit}}$ threshold value. In this case recall drops slightly, which means that some critical time steps are not predicted correctly.

\begin{figure}[!ht]
\centering
\includegraphics[width=0.48\textwidth]{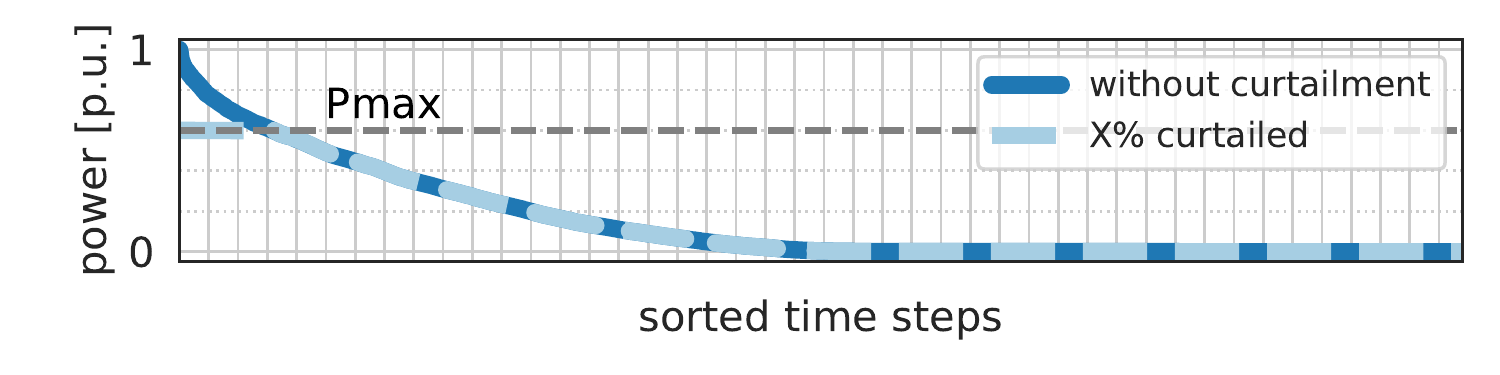}
\caption{Sorted annual curve of the P inputs with and without 3\,\% curtailment. The P inputs without curtailment are used for training; the inputs with 3\,\% curtailment are used for prediction.}
\label{curtailment_inputs}
\end{figure}

We want to assess if a trained regressor/classifier can also predict unknown in-feed situations resulting when curtailing \ac{RES} generation without re-training of the model. As an example, we compute the power flow results for the same year and N-1 cases but with a curtailment of 3\,\% of the energy generated by \ac{RES}. The curtailment of renewable in-feed to reduce investments in the grid infrastructure is suggested by federal law in Germany \cite{eeg2017}. \ac{MLP} models are trained with 10\,\% of the power flow results \textit{without} curtailed generation. We then use the trained \ac{MLP} to predict the critical system states or line loadings based on the curtailed real power values as inputs. Fig.\,\ref{curtailment_inputs} shows a normalised sorted annual curve of these real power values. The real power inputs without curtailment are used for training, where the inputs with 3\,\% curtailment are used for prediction.

\begin{figure}[!ht]
\centering
\includegraphics[width=0.49\textwidth]{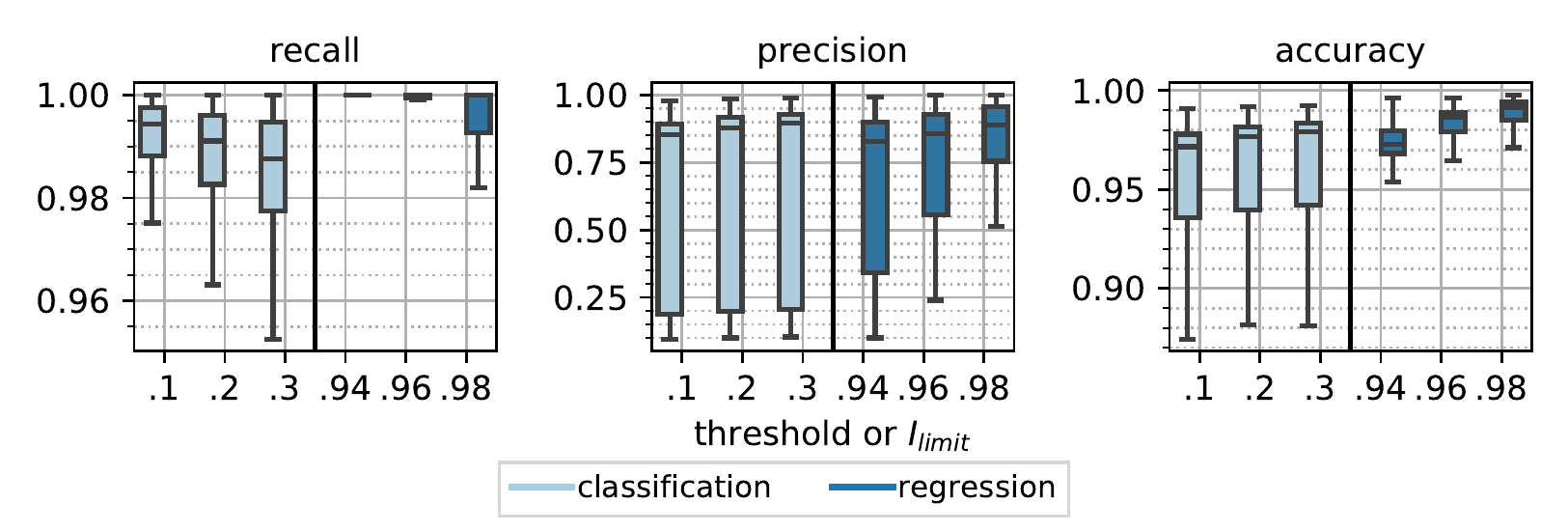}
\caption{Comparison of classification and regression results for the MLP models on untrained data.}
\label{regressor_vs_classifier_curt}
\end{figure}

Fig.\,\ref{regressor_vs_classifier_curt} compares the prediction results when testing with the curtailed time series. Recall increases for the classifier in comparison to previous results (see Fig.\,\ref{regressor_vs_classifier}), since less time steps are critical in the test data set due to the curtailed generation. However, precision and accuracy decrease for the classification of time steps. The classifier anticipates the impact of the curtailed generation only to some extend. The regression method shows better results when predicting the curtailed results, since the performance metrics do not decrease as much.

Detailed comparisons are listed in Table\,\ref{tab_comparison}. The regression method has lower FN and FP values for the exemplary thresholds of 0.2 (classifier) and $0.96 \cdot I_{\mathrm{limit}}$ (regressor). These values result in a higher accuracy, lower false positive rates (FPR)s, and lower false negative rates (FNR)s. The FNR is equal to the share of critical time steps that could not be identified. The FPR is equal to share of mislabelled uncritical time steps and increases the computational time. An additional power flow calculation to validate the prediction is needed for each false positive prediction. Of all critical time steps and N-1 case predictions, between 0.0 - 0.48\,\% are not correctly identified by the regressor. This is about half the amount of the classifier. The FPRs of the regressor are between 2.01 - 3.14\,\% in comparison to 2.41 - 8.14\,\% of the classifier. 

\begin{table}[!ht]
\processtable{Direct comparison of MLP regression and classification results on untrained data. Classification threshold = 0.2, regression threshold = $0.96 \cdot I_{\mathrm{limit}}$. (* = lower is better,  ** = higher is better) \label{tab_comparison}}
{\begin{tabular*}{20pc}{@{\extracolsep{\fill}}l|lll@{}}\toprule
& SB mixed & SB urban  & RTS \\
\midrule
FN classification*       &           3,845 &           5,209 &        460 \\
FN regression*           &           1,986 &           2,967 &          0 \\
FP classification*       &          45,617 &          70,707 &     552,500 \\
FP regression*           &          40,854 &          45,446 &     213,010 \\
correct classification**  &        2,263,178 &        2,657,204 &    6,403,968 \\
correct regression**      &        2,269,800 &        2,684,707 &    6,743,918 \\
total classification    &        2,312,640 &        2,733,120 &    6,956,928 \\
total regression        &        2,312,640 &        2,733,120 &    6,956,928 \\
\midrule
FPR classification*      &       2.41\,\% &       3.13\,\% &   8.14\,\% \\
FPR regression*          &       2.15\,\% &       2.01\,\% &   3.14\,\% \\
FNR classification*      &       0.92\,\% &       1.11\,\% &   0.27\,\% \\
FNR regression*       &       0.48\,\% &       0.63\,\% &   0.00\,\% \\
accuracy classification** &      97.86\,\% &      97.22\,\% &  92.05\,\% \\
accuracy regression**     &      98.15\,\% &      98.23\,\% &  96.94\,\% \\
\botrule
\end{tabular*}}{}
\end{table}

\subsection{Comparison of Timings}
\label{section_timings}
Table \ref{timings} lists the time needed to compute \ac{PF} results, including the base case and N-1 cases without parallel computing. For the SimBench cases, with the 15\,min resolution time series, the power flow calculation times are between 2.29 and 2.81 hours for the used hardware. For RTS, it takes nearly 8 hours to compute these results, due to the higher resolution of the time series (5\,min). The training of the \ac{MLP} takes 10-20\,s for each N-1 case and 11-22\,min in total. Prediction times are much lower with a few hundred milliseconds per N-1 case and 10-20\,s in total. As already shown in Fig.\,\ref{regression_results_training_size} and \ref{training_size_times} the regression as well as the classification method should be trained with at least 10\,\% of power flow results from all time steps and  N-1-cases. In total, the overall time needed for the regression and classification method is dominated by the time needed to compute the training data. The overall time can be reduced by using parallel computing for every N-1 case.

\begin{table}[!ht]
\processtable{Time needed to compute power flow (PF) results for training and prediction times of the \ac{MLP} regressor and classifier.\label{timings}}
{\begin{tabular*}{20pc}{@{\extracolsep{\fill}}l|lll@{}}\toprule
& SB mixed & SB urban & RTS \\
\midrule
$(N+1) \cdot T$ PF results [s] & 8,244 & 10,116 & 28,620 \\
10\,\% of PF results [s] & 822 & 1,014 & 2,863 \\
\midrule
regressor training avg. [s] & 660 & 780 & 660 \\
regressor prediction  avg. [s] & 6.6 & 7.8 & 16.5 \\
\midrule
classifier training  avg. [s] & 660 & 936 & 1,320\\
classifier prediction avg. [s] & 16.5 & 20.4 & 19.8\\
\botrule
\end{tabular*}}{}
\end{table}

\section{Training Data from Scenario Generator}
\label{section_scenario}
\begin{figure}[!ht]
\centering
\includegraphics[width=0.49\textwidth]{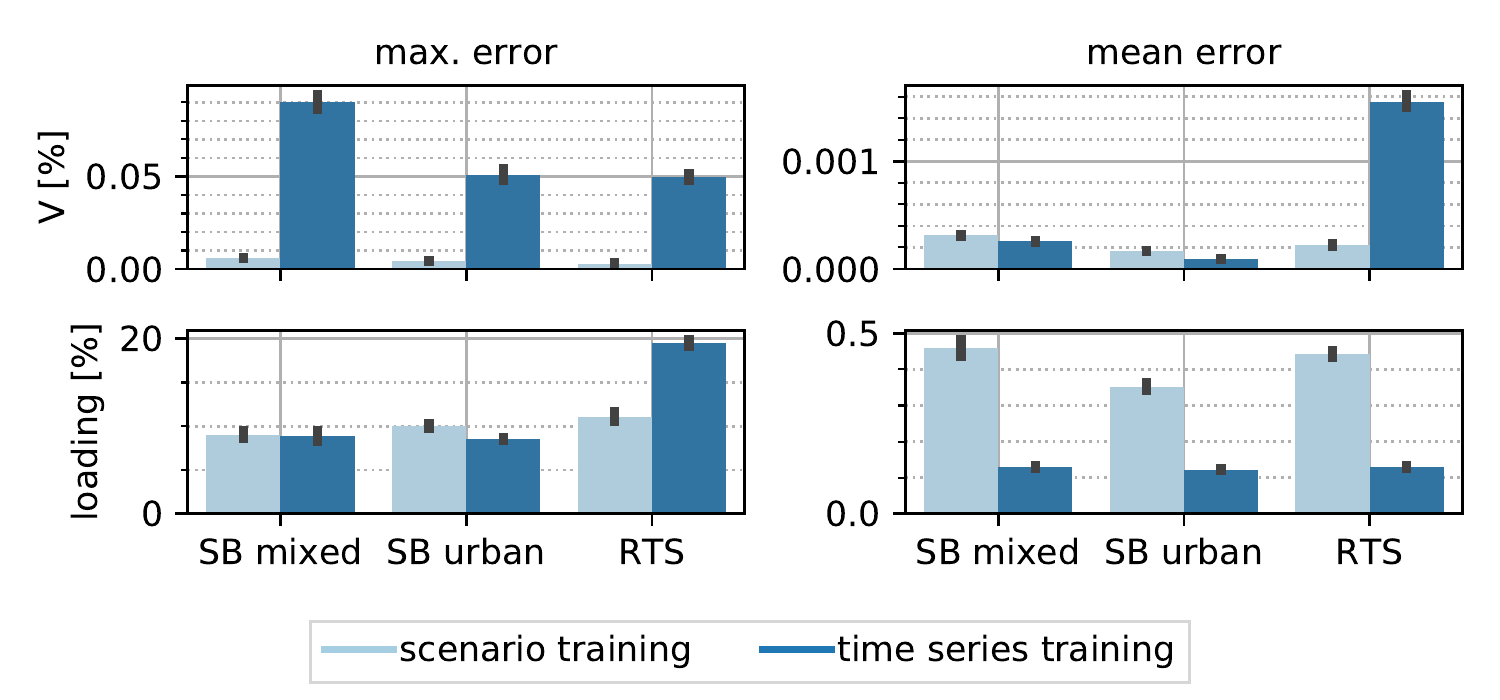}
\caption{Training of the MLP architecture with scenario generator data and from time series data.}
\label{fig_scenario_gen}
\end{figure}

An alternative to using time series for training is to generate training data with a scenario generator, as mentioned in \cite{Menke_2019}. An information-rich data-set enables the ANN to interpolate between the trained scenarios and to estimate the system variables with high accuracy. We consider three different parameters regarding the bus power injections with the scenario generator: the load, RES generator power, and the variation of fossil-fuelled power plant outputs. A scenario consists of a tuple of the scaled values for these three types with ranges between 0\,\% and 100\,\% of their maximum power. The power values are independent of each other. Thus, we scale all units are individually with Gaussian noise to account for variability among the individual units of the same type. We generate the same number of training samples with the scenario generator as we have used for training with the time series data, e.g., 10\,\% of the time series length. Fig.\,\ref{fig_scenario_gen} shows results when using the scenario generator and 10\,\% of the time series results for training. The maximum error of the voltage prediction significantly decreases when using the scenario training for all grids. The maximum line loading prediction error is rather constant for the SimBench grids and decreases only for the RTS case. When regarding the mean errors, the Figure shows that it increases except for the voltage predictions in the RTS case. 

The reduction of the max. errors and the increase of the mean errors can be explained by the similarities of the training data set to the test data set. The majority of samples in the time series training data set is similar to the test data, which is the remaining part of the time series. This similarity results in a low mean prediction error. However, the training data set contains fewer outliers, which increases the maximum error. The scenario generator creates a more balanced training set with fewer outliers that equally distributed. This comparison shows that the ANN architecture generalises well from training data of the scenario generator and that no time series are necessarily needed for training. Such a trained model can be used in live operation to analyse contingency cases in a very short time since the prediction takes only a few milliseconds for each N-1 case. 

\section{Conclusion}
\label{outlook}
We have shown for three test cases that different machine learning algorithms can predict bus voltages and lines loading results. The prediction and training times are much shorter in comparison to the time needed to compute the power flow results. In all comparisons, the MLP architectures have shown the highest prediction accuracy. The XGB classifier has shown good results to identify critical time steps. All other tested regressors and classifiers were not as accurate, did not improve with more training data, or needed much more time to predict results. Training and prediction times for the sklearn \ac{MLP} regressor and classifier were similar. Since the classification of time steps in "critical" and "uncritical" was not faster and also less accurate than the regression method, we recommend using the \ac{MLP} regressor to predict the critical contingency states. Another advantage is that bus voltage magnitudes $V_m$ and line loadings $I_\%$ are outputs of the multivariable prediction instead of binary classification. We noted that the mean and maximum prediction errors decreased with more training data, but also that the majority of time is needed to calculate these training inputs (the power flow results). We found an acceptable trade-off between calculation time and prediction error by selecting 10\,\% of all time steps for training. This resulted in mean errors of 1-2\,\% of line loading, and voltage magnitude predictions for the MLP regressor. The maximum error was in the range between 10-20\,\% for line loading, and around 0.5\,\% for voltage magnitudes. An alternative training method with scenario generator data shows that the MLP can generalise well. 

The use of the prediction method is manifold. In power system planning, the method allows predicting multiple future grid states to evaluate losses or predict contingency situations when integrating \ac{RES}. In live operation, N-1 security states can be assessed in seconds by using the trained \ac{ML} model as a surrogate. High security margins can be considered by using lower prediction thresholds as shown. If multiple future scenarios and thus time series are to be analysed, it might be rather acceptable to have a higher prediction error than longer calculation times since future scenarios are uncertain by definition. The final tolerable error in practice depends on a decision by the grid planner. As a general rule, we recommend considering at least a security margin in the height of the shown maximum errors in live operation as well as in planning. 

The prediction accuracy of \ac{ML} models strongly depends on the available training data. Further research is needed to reduce the number of false negative predictions in imbalanced data sets. Different oversampling and undersampling methods could be tested to reduce the number of these outliers in the training set. Additionally, a combination of training data from the scenario generator and time series could improve the results. For the training of the machine learning algorithms, we used a random train/test split. Since the data is imbalanced and times of high line loadings/voltage magnitudes are correlated with high in-feed, a time step selection based on the input data histogram could increase prediction accuracy. 

\section{Acknowledgments}\label{sec11}
The research is part of the project "SpinAI" and funded by the \ac{BMWI} (funding number 0350030B). The authors are solely responsible for the content of this publication.

\begin{acronym}
\acro{ANN}{artificial neural network}
\acro{AUC}{area under the ROC curve}
\acro{BMWI}{Bundes-Ministerium f{\"u}r Wirtschaft und Energie}
\acro{CAPEX}{capital expenditures}
\acro{DER}{distributed energy resource}
\acro{DG}{distributed generator}
\acro{DSM}{demand side management}
\acro{DSO}{distribution system operator}
\acro{DT}{decision tree}
\acro{GA}{genetic algorithm}
\acro{ICT}{information and communication technology}
\acro{IT}{information technology}
\acro{ML}{machine learning}
\acro{MLP}{multilayer perceptron}
\acro{MILP}{Mixed Integer Linear Programming}
\acro{MINLP}{Mixed Integer Non-Linear Programming}
\acro{OPF}{optimal power flow}
\acro{OPEX}{operating expenditures}
\acro{PF}{power flow}
\acro{PDP}{power distribution planning}
\acro{PSO}{particle swarm optimization}
\acro{RES}{renewable energy sources}
\acro{ROC}{receiver operating characteristics}
\acro{SCADA}{supervisory control and data acquisition}
\acro{SCP}{single contingency policy}
\acro{SMOTE}{Synthetic Minority Over-sampling Technique}
\end{acronym}

\bibliographystyle{iet}
\bibliography{literature}

\end{document}